\documentclass{article}
\usepackage{arxiv}
\usepackage{float}
\usepackage[utf8]{inputenc} % allow utf-8 input
\usepackage[T1]{fontenc}    % use 8-bit T1 fonts
\usepackage{hyperref}       % hyperlinks
\usepackage{url}            % simple URL typesetting
\usepackage{booktabs}       % professional-quality tables
\usepackage{amsfonts}       % blackboard math symbols
\usepackage{nicefrac}       % compact symbols for 1/2, etc.
\usepackage{microtype}      % microtypography
\usepackage{amsmath}
\usepackage{enumitem}
\usepackage[ruled,vlined]{algorithm2e}
\usepackage{lipsum}
\usepackage{graphicx}
\usepackage{natbib}
\graphicspath{ {./images/} }

\title{Conditional Latent Block Model: a Multivariate Time Series Clustering Approach for Autonomous Driving Validation}

\author{
 Etienne Goffinet \\
  Laboratoire Informatique de Paris-Nord\\
  Université Sorbonne Paris Nord\\
  Villetaneuse, France \\
  \texttt{etienne.goffinet@lipn.univ-paris13.fr} \\
   \And
 Anthony Coutant \\
 Laboratoire Informatique de Paris-Nord\\
  Université Sorbonne Paris Nord\\
  Villetaneuse, France \\
  \And
 Mustapha Lebbah \\
  Laboratoire Informatique de Paris-Nord\\
  Université Sorbonne Paris Nord\\
  Villetaneuse, France \\
  \AND
   Hanane Azzag \\
  Laboratoire Informatique de Paris-Nord\\
  Université Sorbonne Paris Nord\\
  Villetaneuse, France \\
  \And
  Loïc Giraldi \\
   Groupe Renault SAS \\
   Avenue du Golf\\
   Guyancourt, France
}

\begin{document}
\maketitle
\begin{abstract}
	Autonomous driving systems validation remains one of the biggest challenges car manufacturers must tackle in order to provide safe driverless cars. The high complexity stems from several factors: the multiplicity of vehicles, embedded systems, use cases, and the very high required level of reliability for the driving system to be at least as safe as a human driver. In order to circumvent these issues, large scale simulations reproducing this huge variety of physical conditions are intensively used to test driverless cars. Therefore, the validation step produces a massive amount of data, including many time-indexed ones, to be processed. In this context, building a structure in the feature space is mandatory to interpret the various scenarios. In this work, we propose a new co-clustering approach adapted to high-dimensional time series analysis, that extends the standard model-based co-clustering. The FunCLBM model extends the recently proposed Functional Latent Block Model and allows to create a dependency structure between row and column clusters. This structured partition acts as a feature selection method, that provides several clustering views of a dataset, while discriminating irrelevant features. In this workflow, times series are projected onto a common interpolated low-dimensional frequency space, which allows to optimize the projection basis. In addition, FunCLBM refines the definition of each latent block by performing block-wise dimension reduction and feature selection. We propose a SEM-Gibbs algorithm to infer this model, as well as a dedicated criterion to select the optimal nested partition. Experiments on both simulated and real-case Renault datasets shows the effectiveness of the proposed tools and the adequacy to our use case.
\end{abstract}

% keywords can be removed

\section{Introduction}

\begin{sloppypar}
	Autonomous car development remains a challenge for car manufacturers. Nowadays, advanced driving assistance systems are being introduced gradually into new car models, yielding more and more complex vehicles that must be proven to be safe. Given the high number of different vehicles, different models, embedded systems, drivers, and expected reliability, physical validation of cars has become prohibitive. \emph{Groupe Renault} has made the technical choice to invest in massive driving simulation technology in order to circumvent this issue. The simulation tool chain mimics car driving conditions based on vehicle physics, driver behavior, and interaction with a configurable environment. The software produces a large quantity of data of excellent quality that needs to be mined.
	The simulation process outputs a large amount of information in the form of multivariate time series. Data size, complexity, and dimensions are considerable: the simulation of a validation test suite, the number of simulations can be as large as $\mathcal{O}(10^6)$, with $\mathcal{O}(10^3)$ signals, each recording $\mathcal{O}(10^4)$ time steps. Overall, this setting implies the production of more than $\mathcal{O}(10^{13})$ data points.

\begin{figure}[h!]
    \centering
    \includegraphics[width=1.0\linewidth]{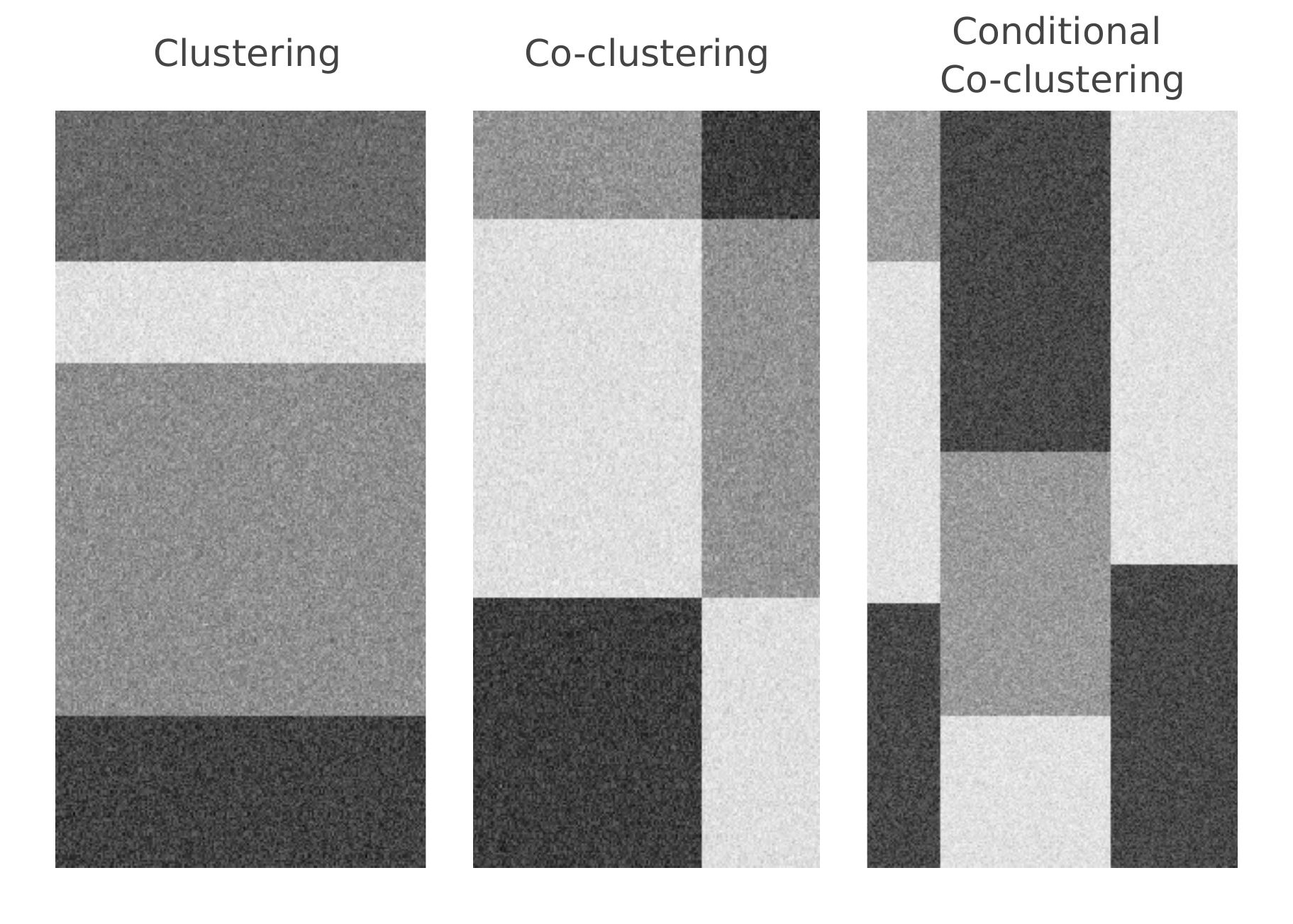}
    \caption{Differences between Mixture Model clustering, Latent Block Model Co-clustering and Conditional Latent Block Model Co-clustering}
    \label{fig:mmvslbmvsclbm}
    \end{figure}

    One issue driving system developers are facing with this large amount of data, is the ability to identify operational modes of the driving systems, in order to better understand and refine the control logic. Specific visualization methods are required for this purpose. Clustering is a first approach to tackle the problem, which consists in the automatic grouping of "similar" observations into homogeneous groups (clusters). The clustering of time series (also called \emph{functional} data) already helps decision-making in many domains (Health, Finance, Industry$\dots$) and has been intensively studied (see \cite{bagnall2017great, aghabozorgi2015time} for reviews). In such methods, observation clusters construction is based on every functional features (see Fig.~\ref{fig:mmvslbmvsclbm}, left panel). 
	
	Co-clustering techniques produce joint clusters of observations and clusters of features. The Latent Block Model (LBM) is a model-based approach to co-clustering which has recently proved its effectiveness in various applications (\cite{govaert2013co,jacques2018model}). It can be applied when every feature can be modeled with the same probability density function, for instance applied to the clustering of text based on word counts, or in the functional case like in \cite{bouveyron2018functional} where features (also called \emph{signals}) come from the day-by-day segmentation of electricity consumption curves. Latent Block Model applied to time series is a recent approach, and there exists only few works on the topic \cite{chamroukhi2017model,slimen2018model,schmutz2019co}. In particular, \cite{bouveyron2018functional} presents an interesting Functional Latent Block Model (\emph{FunLBM}) that relies on functional PCA projections of the series expressed in a Fourier basis. 
	Co-clustering methods enable grouping of similar features with, in our applications, a limiting constraint: observation clusters and feature clusters are independent and observation partition is common to every features (see Fig.~\ref{fig:mmvslbmvsclbm}, middle panel). 
	
	In real cases, chances are that for every feature cluster there is a different set of observation clusters. The main advantage of this new model, called Functional Conditional Latent Block Model (\emph{FunCLBM}), is that a joint structure is introduced in the clusters dependency: clusters are not assumed to be independent anymore and observation clusters depend on feature ones. Consequently, users have at disposal a clustering made of multiple views, from which it is easy to discard groups of useless features. This construction grants a valuable tool to the expert: feature selection and discrimination of uninformative features. Fig.~\ref{fig:mmvslbmvsclbm} shows the differences between the clustering, co-clustering, and the proposed conditional co-clustering approaches. 
	
	In the most recent FunLBM works, time series are expressed in a common polynomial basis and block-wise functional PCAs (\cite{ramsay2005principal}) are applied on the regression coefficients. FunCLBM builds on this construction and improves the first part: the time series are first transformed with a Discrete Fourier Transform procedure, then obtained periodograms are interpolated in order to construct a finely-tuned expression basis.
	The rest of this paper is organized as follows: Section 2 presents the work related to both model-based clustering and co-clustering. Both data processing aspects and the FunCLBM workflow are detailed in Section 3. Section 4 describes the inference and implementation details. Experiments on both a simulated dataset and a real case data are presented in Section 5. Finally, the paper ends on Section 6 with future work perspectives.

\section{Related work}

\subsection{Model-based clustering} \label{MM}

Mixture modeling (\emph{MM} is a standard clustering approach first introduced in \cite{dempster1977maximum} and based on the assumption of latent clusters. The cluster membership probabilities are jointly estimated with the mixture parameters: the proportions and the distribution parameters of each component. In opposition to non-model-based methods, this approach enables the construction of confidence intervals and probabilistic outliers detection.
The inference is performed by optimizing the likelihood of the model, with a dedicated algorithm, the Expectation-Maximization (EM) algorithm \cite{dempster1977maximum}. This algorithm is iterative and composed of two steps: the E step computes the posterior cluster membership probabilities while the M step updates the model parameters based on these probabilities.

Several variants of this algorithm exist, which mainly consist in variations of the M step. The Classification-EM version updates the parameters based on the observations that are "most likely" to belong to each cluster (this version is the closest to the popular K-Means, of which it can be considered a probabilistic generalisation). In the Stochatisc-EM (SEM) approach, the cluster belongings are drawn at random according to the membership probabilities. FunCLBM uses this last EM extension, in an adapted version detailed in Section 4. 
Model-based clustering has been subject to many works, improving several aspects (e.g. the initialization \cite{biernacki2003choosing} or the model selection strategy \cite{vlassis2002greedy,biernacki2000assessing}). It has also been extended to the modeling of various data types, including time series \cite{bouveyron2011model, chamroukhi2010hidden}
The Latent Block Model is an extension of the MM model addressing the co-clustering problem.

\subsection{Model-based co-clustering}

The Latent Block Model, proposed in \cite{govaert2013co}, assumes the presence of latent feature clusters (\emph{column-cluster}) in addition to the observation clusters (\emph{row-cluster}). In the standard multivariate MM, each component of the mixture defines a density over multivariate observations (an observation corresponding to a row). It is no longer the case in the LBM framework: the modeled object is the \emph{cell}, i.e the intersection of a row and a column, that is an observation for a given feature.

Inside the block component created by crossing the column and row partition, LBM assumes the independence conditional to the block, which means that, given a block partition, every cell composing a block is independent of each other. From this perspective, the density of an LBM component is univariate. 

Given an observed dataset $x=(x_{i j})_{n\times p}$ of $n$ observations of $p$ features, let denote $z=(z_{i k})_{n\times K}$ and $w=(w_{j \ell})_{p\times L}$ the random binary matrices indicating respectively the row and cluster partitions. The standard LBM is defined by:
\begin{equation*}
\mathrm{p}(x;\theta)= \sum_{\mathcal{Z}\times\mathcal{W}} \mathrm{p}(z;\theta) \mathrm{p}(w;\theta) \mathrm{p}(x|z,w;\theta),
\end{equation*}
where $\mathcal{Z}$ and $\mathcal{W}$ respectively denote the sets of all possible row and column partitions. The quantity $\mathrm{p}(z;\theta)$ is defined as $\prod_{i k} \pi_k^{z_{i k}}$, with $\pi=(\pi_k)_K$ the membership probabilities prior (respectively for $w$ with the membership probabilities $\rho=(\rho_\ell)_L$). The set of parameters $\theta$ is composed of the mixing proportions and of the component density parameters. These densities $\mathrm{p}(x|z,w;\theta)$ are part of a common model family suited to clustering interpretation.

We emphasize that our goal is to provide a cluster belonging probability that is not independent anymore, i.e. the approximation $\mathrm{p}(z,w;\theta)\approx\mathrm{p}(z;\theta)\mathrm{p}(w;\theta)$ does not hold.

\subsection{Functional co-clustering}

In the case of time series co-clustering, the dataset is composed of sequences: $x=(x_{i j})_{n\times p}$, with $x_{i j}= (x_{i j}(t))_{T}$ and $T$ the time support. Each time series model-based co-clustering method defines and makes use of a specific representation of the time series and probability density functions for the mixture components. While one article from \cite{chamroukhi2017model} uses a density based on a piecewise regression model, the majority of them \cite{slimen2018model,bouveyron2018functional, schmutz2019co} are based on modeling the time series using a functional PCA (fPCA) projection \cite{ramsay2005principal}. This process assumes the dataset time series can be adequately represented in a common low-space expansion basis, i.e. each $x_{i j}$ can be expressed as $x_{i j}(t)=\sum_{s=1}^S c_{i j s} f_s(t)$. This projection allows to reconstruct the functional form from the discrete time series representation. Fourier basis is a common choice in the domain. 

\begin{figure}[h!]
	\centering
	\includegraphics[width=0.8\linewidth]{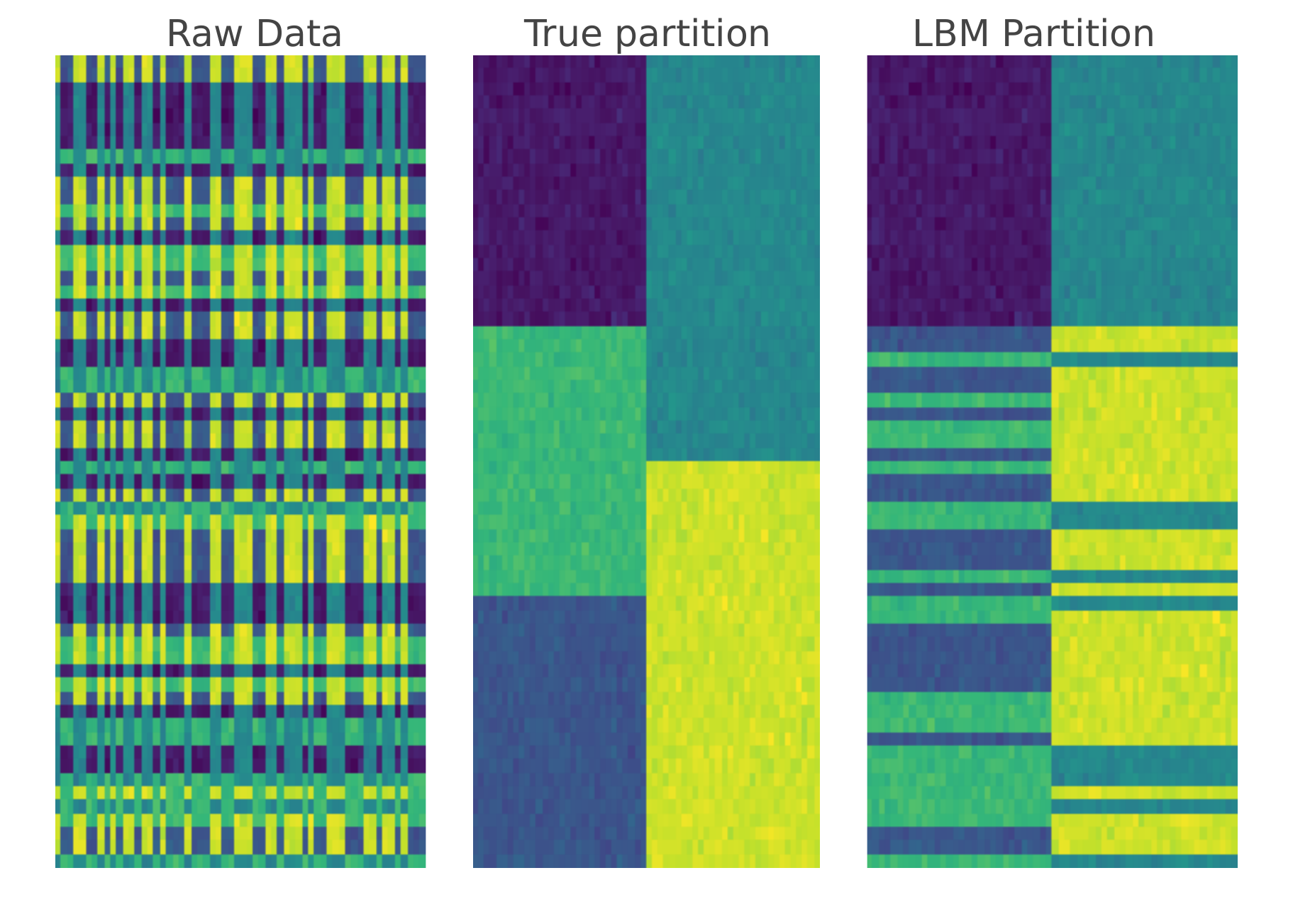}
	\caption{Several reordered views of a conditionally partitioned univariate Gaussian dataset: the original "random" view, the optimal "true" partition and the best partition that can be produced with a standard Gaussian LBM approach}
	\label{fig:lbmvsclbm}
\end{figure}

The LBM is then applied to the coefficients dataset. Applying the fPCA block-per-block, as presented in  \cite{bouveyron2018functional}, allows to detect even the smallest signal change. This work shows good performances but cannot deal with datasets that contain irrelevant, uninformative features or in the case column clusters define different row clusters. In this situation, using the LBM forces to make compromises in the block partitioning, resulting in sub-optimal block clustering solutions. Fig.~\ref{fig:lbmvsclbm} illustrates this behavior with an example of univariate Gaussian Latent Block Model co-clustering. FunCLBM allows to extend the FunLBM in order to overcome this limitation.

\section{Functional Conditional Latent Block Model}

This section presents the FunCLBM model, as well as its inference and model selection strategies. The proposed approach relies on the projection of the time series in a specific space. It consists in applying a PCA on the representation of the series in the frequency domain.

\subsection{Representation with interpolated Fourier transform}

This paper focuses on working with high-dimensional time series. Using such series as is makes learning hard due to both the \emph{curse of dimension}, tending to make all individuals equally distant from each other, and the huge quantity of noise involved.
It is required to choose a more compact representation for learning. Many have been studied in the literature (Fourier, wavelets, Chebyshev, \ldots) each with its pros and cons.
In this paper, an interpolated log-scaled Fourier periodogram representation is chosen, following what is advocated in \cite{caiado2009comparison}.
Formally, given a family of series $X = (x_{i j})_{n\times p}$, each $x_{i j}$ with its own time length $l_{i j} = |x_{i j}|$ (its number of discrete time points), the first step is to compute the Fourier periodograms $P = (p_{i j})_{n\times p}$ of same length. These periodograms are not indexed over time though but over frequencies $F = (f_{i j})_{n\times p} = (\langle f_{i j}^1, \ldots, f_{i j}^{l_{i j}}\rangle )_{n\times p}$. 

However, the different length of each time series makes the discrete periodogram frequencies unaligned, so that the $f_{i j}$ values for 
a dataset of series are likely to be significantly different between series. Thus, in order to make the representation
comparable, one needs to find a common sequence of frequencies $\widehat{f}$ from $F$ which is not 
obtainable by selecting a subset of each series frequencies. 
A possible solution chosen for this paper is to obtain $\widehat{f}$ by computing the sample average gap $\widehat{\Delta f}$ between two consecutive frequencies over all time series.
Then, choosing a desired representation length $\widehat{l}$, it is possible to build $\widehat{f} = \langle 0, \widehat{\Delta f}, 2 \widehat{\Delta f}, \ldots, (\widehat{l} - 1) \widehat{\Delta f} \rangle$.

The final step is to obtain the periodogram values $\widehat{P} = (\widehat{p}_{i j})_{n\times p}$ of all series of $X$ for $\widehat{f}$ frequencies, 
which can only be estimated, e.g. using linear or cubic interpolation techniques from $P$.
One could be tempted to use $\widehat{P}$ directly as the compact representation for model selection. 
However, as advised in \cite{caiado2009comparison}, $\text{log}\left(z\left(\widehat{P}\right)\right) = (\text{log}\left(z\left(\hat{p}_{i j}\right)\right)_{n\times p}$, 
where $z$ is the z-normalization function substracting the mean value and dividing by the standard deviation, 
is used instead to compare relative periodogram values and limit the bias towards low frequencies  encountered in practice. 

\subsection{Model definition}

Considering independent row-clusters partition for each column-cluster, it is mandatory to adapt the previous notation. Let denote $K_\ell$ the number of row-clusters associated to column-cluster $\ell$, $1\leq \ell \leq L$. We still denote by $w=(w_{j \ell})_{p\times L}$ the binary vector that indicates the column-cluster belonging. Given a column-cluster $\ell$, the associated row-clusters partition is denoted as $z^\ell=(z_{i k}^\ell)_{n\times K_\ell}$. For convenience, we note $z_i^\ell=(z_{i k}^\ell)_{K_\ell}$ the row-cluster membership of observation $i$ in the column cluster $\ell$. The global row-partition is denoted as $z=(z_\ell)_L$. While the column mixing proportion remains the same, the row mixing proportion is now denoted $\pi=(\pi_\ell)_L$, with $\pi_\ell=(\pi_{k_\ell})_{K_\ell}$. Finally, the joint model density can be decomposed in the following:
\begin{align*}
\mathrm{p}(x;\theta) 
&= \sum_{\mathcal{Z}\times\mathcal{W}} \mathrm{p}(w;\theta) \mathrm{p}(z|w;\theta) \mathrm{p}(c|z,w;\theta) \\
&= \sum_{\mathcal{Z}\times\mathcal{W}} \prod_{j \ell} \rho_l^{w_{j \ell}} \prod_{i \ell} \prod_{k_\ell} {\pi_{k_\ell}^\ell}^{z_{i k}^l} \prod_{i j \ell} \prod_{k_\ell} \mathrm{p}(c_{i j}; \theta_k^\ell)^{z_{i k}^\ell w_{j l}},
\end{align*}
with $\mathrm{p}(c_{i j}; \theta_k^\ell)=\mathrm{p}(c_{i j}|w_j, z_i^l)$ being the density of the block $(k_\ell,\ell)$ with parameters $\theta_k^\ell$. This density is the one of a multivariate gaussian model on the projections of the interpolated periodogram coefficients into a low-dimensional subspace. This density is parameterized by three elements: 
\begin{itemize}
	\item A matrix $A_k^\ell$ of size $m\times d$ which defines the linear transformation of the periodogram (of size $m$) in the lower-dimension subspace (of size $d$).
	\item The m-dimension mode $\mu_k^\ell$.
	\item $\Sigma_k^\ell$, the $d\times d$ covariance matrix in that subspace.
\end{itemize}

The complete set of parameter $\theta$=$(\pi,\rho, (\theta_k^\ell)_{K_\ell\times L})$ is inferred with a dedicated SEM-Gibbs algorithm.

\subsection{Inference with SEM-Gibbs algorithm}

Using the SEM algorithm is a popular practice in the model-based clustering framework. As described in section \ref{MM}, the SEM implies that the block component parameters are updated based on sampled observations. In the co-clustering case, there is an additional constraint: the direct computation of the block belonging is intractable. A popular solution from \cite{keribin2010estimation} is to use a Gibbs sampler that alternatively draws the cluster belongings in one dimension conditionally to the other. Starting from an initial parameter state $\theta^0$ and an initial column partition $w^0$, the SEM-Gibbs alternates between these two steps:

\begin{enumerate}
	\item SE step: 
	\begin{itemize}
	    \item For each column partition $\ell$ and each row $i$, draw the associated row cluster belonging ${z_i^\ell}^{(q+1)} \sim \mathcal{M}(1,\tilde{z}_{i 1}^\ell,\dots,\tilde{z}_{i K_\ell}^\ell)$, with
    	\begin{align*}
    	    \tilde{z}_{i k}^l &= \mathrm{p}\left(z_{i k}^\ell=1 | c_{i .}, w^{(q)} ; \theta^{(q)}\right) \\ 
    	    &=\frac{{\pi_k^\ell}^{(q)} {f_k^\ell}\left(c_{i .}| w^{(q)} ; \theta^{(q)}\right)}{\sum_{h=1}^{K_\ell} {\pi_h^\ell}^{(q)}  {f_h^\ell}\left(c_{i .} | w^{(q)} ; \theta^{(q)}\right)},
	    \end{align*}
	    
	    where $c_{i .}=(c_{i j})_{0\leq j\leq p}$ and ${f_k^\ell}$ the density of the row: \\
	    ${f_k^\ell}\left(c_{i .} | w^{(q)} ; \theta^{(q)}\right)=\prod_{j} \mathrm{p}\left(c_{i j} ; \theta_{k \ell}^{(q)}\right)^{w_{j \ell}^{(q)}}$.
	    \item For each column $j$, draw the column cluster belonging $w_j^{(q+1)} \sim \mathcal{M}(1,\tilde{w}_{j 1},\dots, \tilde{w}_{j L})$, with
	    \begin{align*}
    	    \tilde{w}_{j \ell}&=\mathrm{p}\left(w_{j \ell}=1 | c_{. j}, z^{(q+1)} ; \theta^{(q)}\right)\\
    	    &=\frac{\rho_\ell^{(q)} g_{\ell}\left(c_{. j} | z^{(q+1)} ; \theta^{(q)}\right)}{\sum_{r=1}^L \rho_{r}^{(q)} f_r\left(c_{. j} | z^{(q+1)} ; \theta^{(q)}\right)},
	    \end{align*}
	     where $c_{. j}=(c_{i j})_{0\leq i\leq n}$ and $g_{\ell}$ is the density of the column $c(.j)$ given the multiple row partition:
	     \begin{align*}
    	     g_{\ell}\left(c_{. j} | z^{(q+1)} ; \theta^{(q)}\right)=\prod_{i k} \mathrm{p}\left(c_{i j} ; \theta_{k \ell}^{(q)}\right)^{{z_{i k}^l}^{(q+1)}}
	     \end{align*}
	\end{itemize} 
	
	\item M Step: given the sampled block partition, and denoting by $c_k^\ell$ the observations belonging to block $(k,\ell)$, the mixture proportions are updated by:
	\begin{itemize}
		\item ${\pi_{k_\ell}}^{(q+1)}=\frac{1}{n}\sum_{i} {z_{i k}^l}^{(q+1)}$, $0 \leq \ell \leq L$,
	    \item  ${\rho_\ell}^{(q+1)}=\frac{1}{p}\sum_{j} {w_{j \ell}}^{(q+1)}$
	    \item $A_k^{\ell}$ the loadings matrix produced by the block-wise PCA of $c_k^\ell$, i.e. the $m\times d$ matrix containing the $d$ eigenvectors with highest eigenvalues.
	    \item $\mu_k^{\ell}$ and $\Sigma_k^{\ell}$ the mean and covariance matrices in the lower-dimensional subspace:
	   \begin{gather*}
            \mu_k^\ell= \frac{1}{{n_k^ \ell}^{(q+1)}} \sum_{i,j} {z_{i k}^\ell}^{(q+1)} w_{j l}^{(q+1)} v_{i j} \\
            \Sigma_k^\ell= \frac{1}{{n_k^ \ell}^{(q+1)}}\sum_{i,j} {{z_{i k}^\ell}^{(q+1)} w_{j l}^{(q+1)} \left(v_{i j} - \mu_k^\ell\right)\left(v_{i j} - \mu_k^\ell\right)^T},
        \end{gather*}
	    with $v_{i j}=c_{i j} A_k^\ell$, and ${n_k^ \ell}^{(q+1)}=\sum_{i} \sum_{j} z_{i k}^{(q+1)} w_{j \ell}^{(q+1)}$
	\end{itemize}
\end{enumerate}

This algorithm is run for a given number of iterations, or until a relative convergence threshold is reached.
Choosing a good initialization state is crucial in order to ensure the good behaviour of the algorithm. It is a well-known dilemma in model-based clustering, subject of several works \cite{blomer2013simple,baudry2015mixtures}. 

Several methods are often considered: populating components with a small random sample of the observations, shuffling the column and block partitions, or using another clustering algorithm to get a good initial starting point. In section 4 these different initializations are experimented. Independently from the method, taking the result with highest likelihood among several runs is an agreed-upon strategy. 

\subsection{Model Selection}

With a good initialization choice, SEM-Gibbs may converge to a solution for a given clustering structure, i.e. a column cluster number $L$ and a set of row clusters numbers $K=(K_\ell)_L$. Several criteria have been developed to address the model selection problem. In this work, we propose a dedicated criterion based on the Integrated
Classification Likelihood (ICL) \cite{biernacki2000assessing}. Initially developed for Gaussian Mixture Model Selection, extended by \cite{lomet2012selection} to co-clustering and in \cite{bouveyron2018functional} to functional co-clustering, we propose the following extension to functional conditional co-clustering:
\begin{align*}
    \operatorname{ICL}(K, L)&=\log \mathrm{p}(\boldsymbol{x}, \hat{\boldsymbol{v}}, \hat{\boldsymbol{w}} ; \hat{\theta})-\frac{L-1}{2} \log p  \\
    &-\frac{1}{2}\sum_\ell ((K_\ell-1) \log n)-\frac{\sum_{\ell, k} \nu_k^\ell}{2} \log (n p),
\end{align*}
where $\nu_k^\ell$ is the component parameter number of block $(k,\ell)$. This score penalizes the log-likelihood with a function of the number of parameters. The best model is the one maximizing this score. In the co-clustering case, finding the best structure can be done by an exhaustive grid search, we will see in the experiments that this strategy is not suitable to FunCLBM.

\section{Experiments on Synthetic Data}

In order to test the capabilities of FunCLBM, experiments are first conducted on a simulated dataset. These experiments help us check that the model is suited and that the SEM-Gibbs algorithm behaves well in a controlled environment.

\subsection{Simulated dataset}

\begin{figure}[h!]
	\centering
	\includegraphics[width=0.7\linewidth]{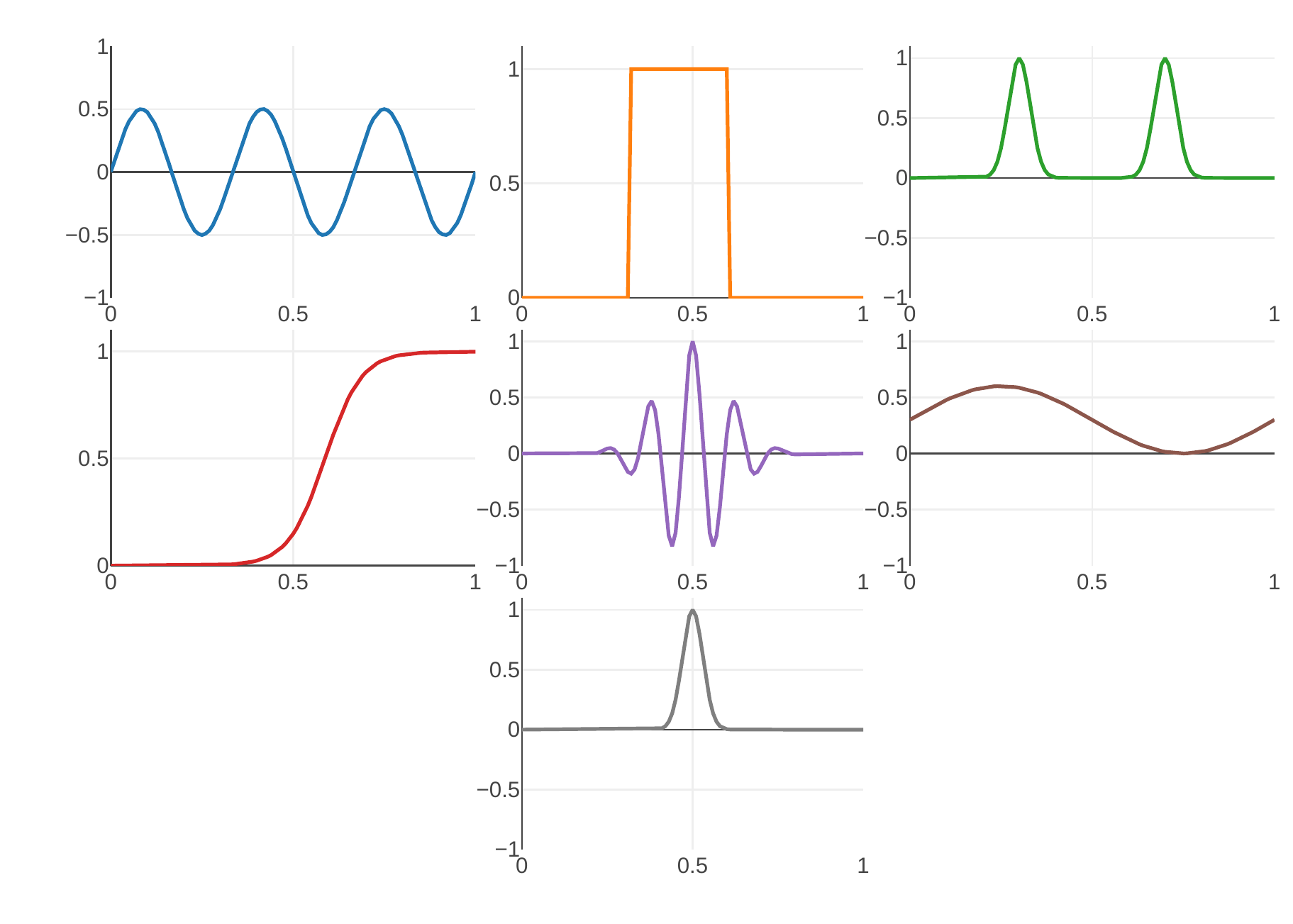}
	\caption{Prototypes used as block mode for the simulations}
	\label{fig:prototypes}
\end{figure}

The first experiment is conducted on a dataset sampled from a known generative model. The objective is to check the behavior of FunCLBM, its initialization and model selection strategies. The dataset is generated by sampling around one of several "prototypes" denoted $(\phi_k^\ell)$ and which represents the components modes in the original space. For each block $(k,\ell)$, several time series are drawn following $\mathcal{N}\left(\phi_{k \ell}(t+t_s), s^{2}\right)$ with $s= 0.02$ and $t_s$ a random shift $\sim \mathcal{N}\left(0, s^{2}\right)$. These modes are depicted in Fig.~\ref{fig:prototypes} according to the dataset structure.

In the experiments, the quality of the estimated block partition is compared to the known generative partition, based on the Adjusted Rand Index (ARI). This is a popular criterion choice in the clustering domain, which represents the proportion of correctly grouped and separated observations with respect to the observed classes.
In our particular context, we compare the obtained partition based on three aspects: the column cluster partition, the rows cluster partitions (made of the binning of every row cluster partition per column), and the block partition. We generate a dataset of size 90x90, with column cluster of size $(45, 15, 30)$ and row cluster sizes of respective sizes $(20, 40, 30)$, $(60, 30)$ and $(40, 50)$.

\subsection{Model Adequacy}

\begin{figure}[h!]
	\centering
	\includegraphics[width=0.7\linewidth]{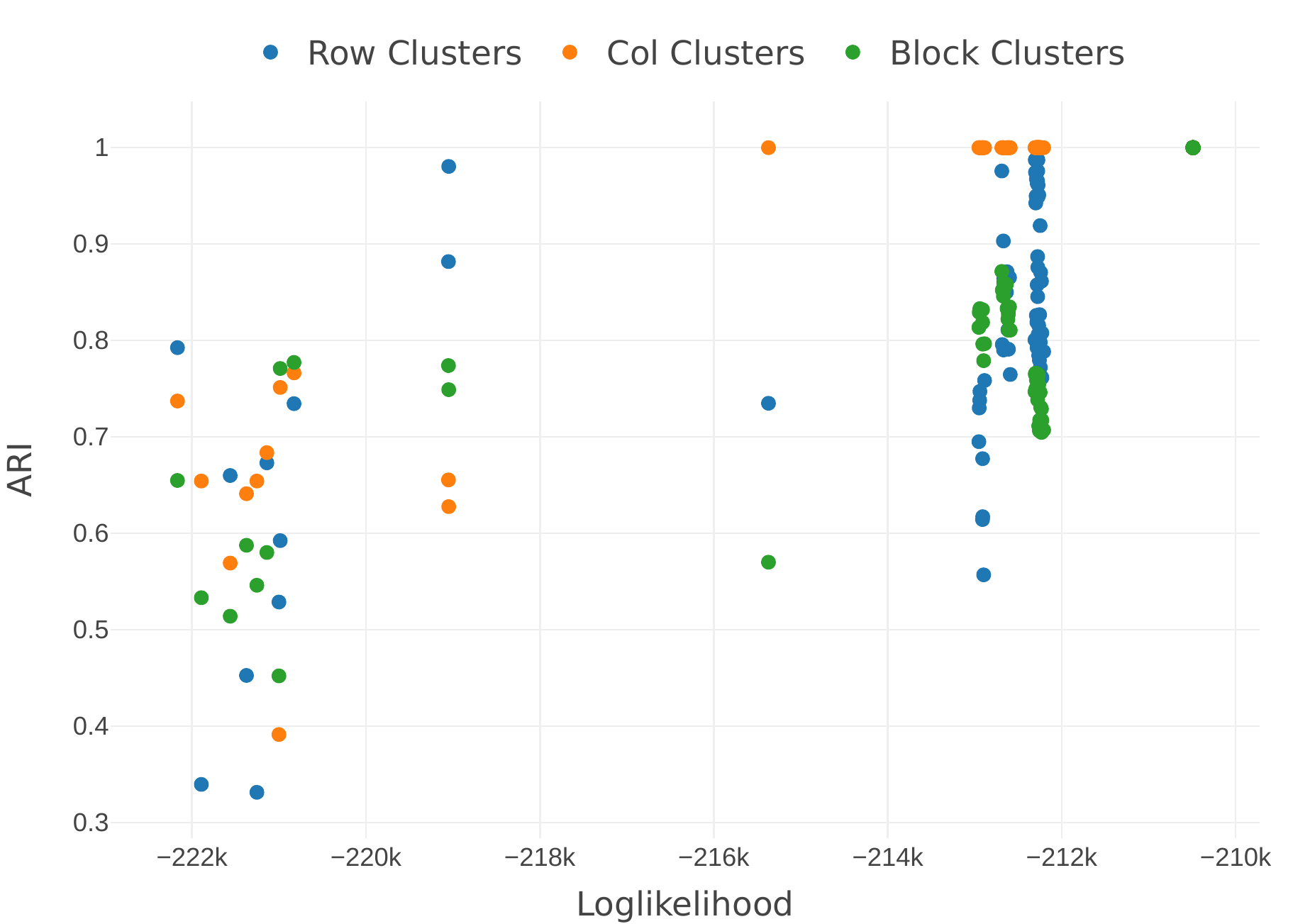}
	\caption{ARI versus Log-Likelihood in 100 launches of SEM-Gibbs on the simulated dataset}
	\label{fig:aricsvll}
\end{figure}

As a preliminary test, we verify that FunCLBM objective function in lower-dimensional spaces is suited to the clustering of such dataset. To do so, we compare the Log-likelihood produced through 100 launches of SEM-Gibbs to the corresponding ARI criterion and depicted in Fig.~\ref{fig:aricsvll}. We verify this relationship by computing the Pearson's correlation coefficient between the two scores, and Kendall's correlation test. We use this latter test to avoid making assumptions on ARI or Log-likelihood normality and because of the presence of ex-aequos values that can be produced if the "true state" is reached. The test results (with 95 $\%$ confidence level) are displayed in Table~\ref{tab:pvaluesCorrelationARILL}. For this dataset, the suitability of the method is attested by the strong Pearson's correlation for every partition dimension and confirmed by Kendall's correlation test p-value at 95\% confidence level.

\begin{table}[h!]
    \caption{Kendall's correlation test p-value (conf level: 0.95)}
     \centering
    \begin{tabular}{|c||c|c|c|}
    \hline  & Row &  Column &  Block \\
    \hline  Pearson's correlation & 0.7110134 & 0.8974437 & 0.7111690  \\
    \hline  Kendall test p-value & 7.88e-18 &  6.091e-08 & 8.09e-06  \\
    \hline 
    \end{tabular}

    \label{tab:pvaluesCorrelationARILL}
\end{table}

\subsection{Initialization}

The next experiments aim at evaluating the different initialization strategies. It is, once more, evaluated with the ARI criterion. We compare four strategies:

\begin{itemize}
    \item Populate blocks with samples
    \item Random shuffle of the column partition, then of the row partition
    \item Initialize the column partition with a K-Means run on the transposed dataset, then the row partition with one K-Means run per column.
    \item Initialize the column partition with a model-based functional co-clustering approach.
\end{itemize}

\begin{figure}[h!]
	\centering
	\includegraphics[width=0.7\linewidth]{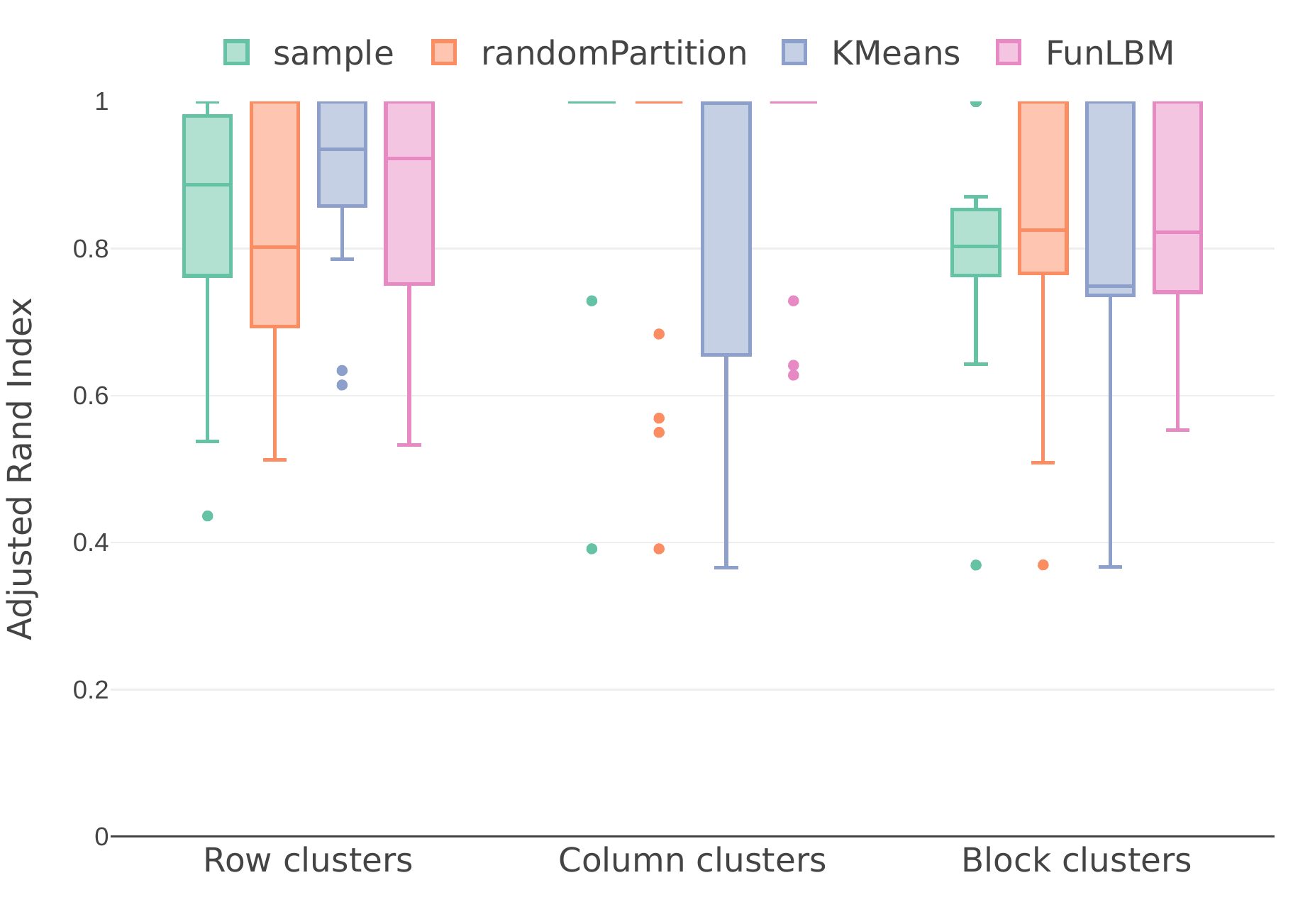}
	\caption{Results of Row, Cluster and Block partition ARI obtained with different initialization methods (median and quantile 0.9 on 30 SEM-Gibbs runs)}
	\label{fig:ariVsInitialization_exp4}
\end{figure}

For the last approach, we use a modified version of the original FunLBM, closer to the FunCLBM variant: the time series are transformed into interpolated log-periodograms and component parameter updates are the same as in FunCLBM (c.f. Section 3). 

In Fig.~\ref{fig:ariVsInitialization_exp4} are displayed the ARI obtained after launching 30 times SEM-Gibbs with each initialization method. The figure shows important differences between row and column cluster results, as expected since the model does not treat rows and column symmetrically anymore. We can also observe that the K-Means run has an unexpected behavior: while performing well on average, its results show a high dispersion for column cluster ARI. Its row cluster ARI however is slightly better than the other methods. On average, the model-based co-clustering performs well but not overwhelmingly. 

\begin{figure}[h!]
	\centering
	\includegraphics[width=0.7\linewidth]{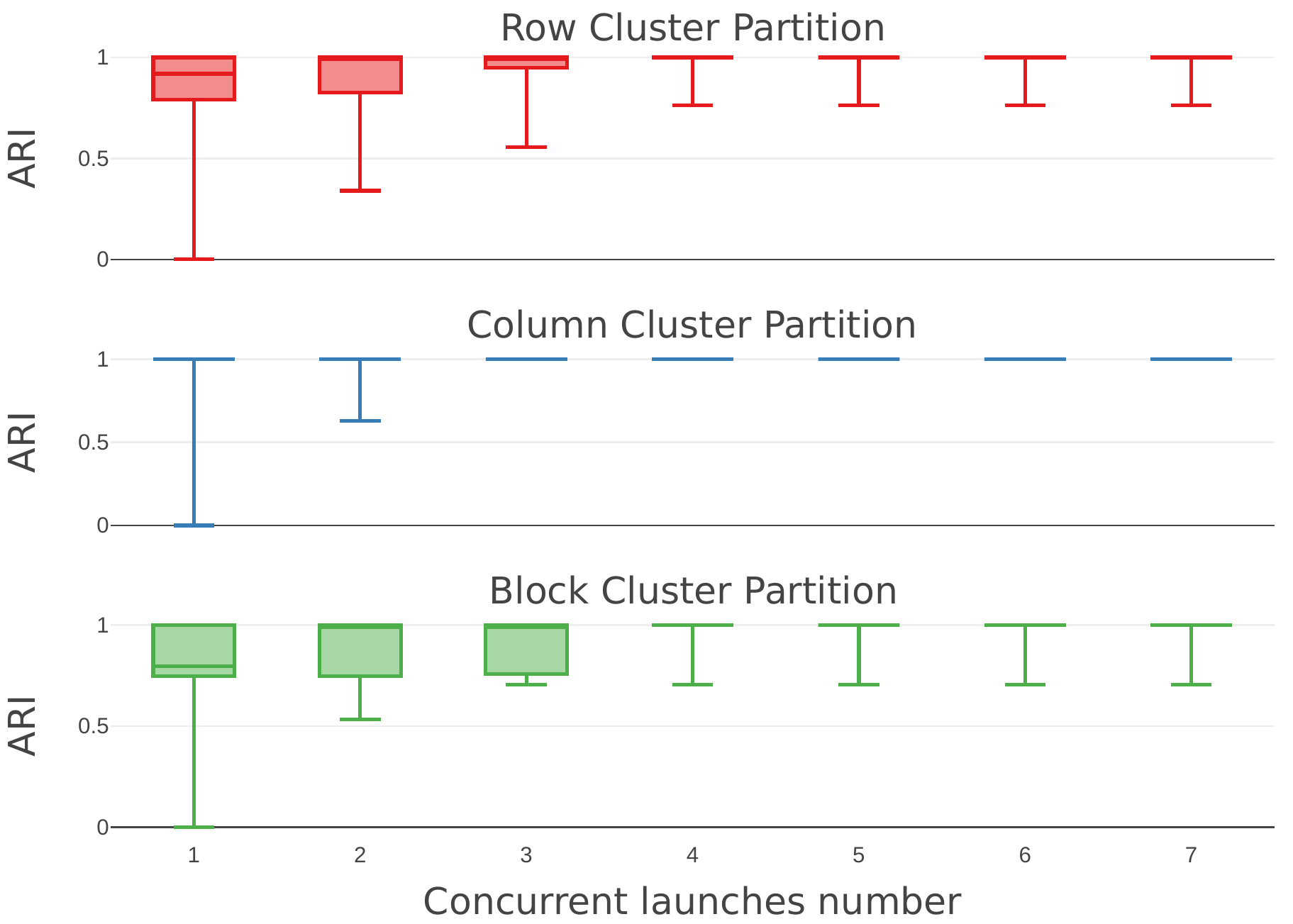}
	\caption{Best ARI obtained among several SEM-Gibbs runs (median, quantile 0.9; with variable number of concurrents)}
	\label{fig:stability}
\end{figure}

On this small experimental case, the random partition seems a direct and cheap initialization strategy. Whichever initialization strategy is applied, the concurrent run of several methods allows to stabilize the results, as displayed in Fig.~\ref{fig:stability}. 
In this experimental setup, finding the perfect structure is an easy task whenever the number of concurrent launches is higher than 4.

\subsection{Model selection}

The last experiments compared the initialization methods for a given choice of structure. However, the most challenging part, and also the most useful for the field expert, is the model selection strategy. In \cite{bouveyron2018functional}, the authors proposed the co-clustering grid search method, which requires inferring $K\times L$ models. In the clustering case, the number of components is preferably low for interpretability sake. In our case, such approach is not possible: the number of combinations is prohibitive. For a maximal number of column clusters $L_{M}$ and per-column row clusters $K_{M}$, it is the number of un-ordered set of length $\ell$ among $K_{M}$ possibilities, for $0 \leq \ell \leq L_{M}$, i.e. 
$\sum_{\ell=1}^{L_{M}} \binom{K_{M}+\ell -1}{\ell}$. The quantity is prohibitive: with $L_M=5$, $K_M=5$,
it amounts to 251 combinations, i.e. 10 times more than in the LBM case (5x5 combinations).

In order to overcome this limitation, we propose and compare two strategies. Both are based on a different estimation of $\widehat{L}$ and then on a column-wise grid search. 
In the first case, $\widehat{L}$ is estimated from a standard model-based co-clustering exhaustive grid search, and in the second from a greedy algorithm. The first solution implies an exhaustive search of the $K_{M} \times L_{M}$ combinations and then $\widehat{L}\times K_{M}$ to produce the best number of row-clusters per column. The second one is an iterative algorithm that chooses, at each iteration, the best functional Latent Block Model between the one with an added row cluster and the one with an added column cluster. The number of inferences to perform depends on the number of iterations, with upper bound $L_{M} + K_{M}$. After each construction, the candidate is finally taken as initial FunCLBM model state for a new SEM-Gibbs run.

\begin{figure}
	\centering
	\includegraphics[width=0.7\linewidth]{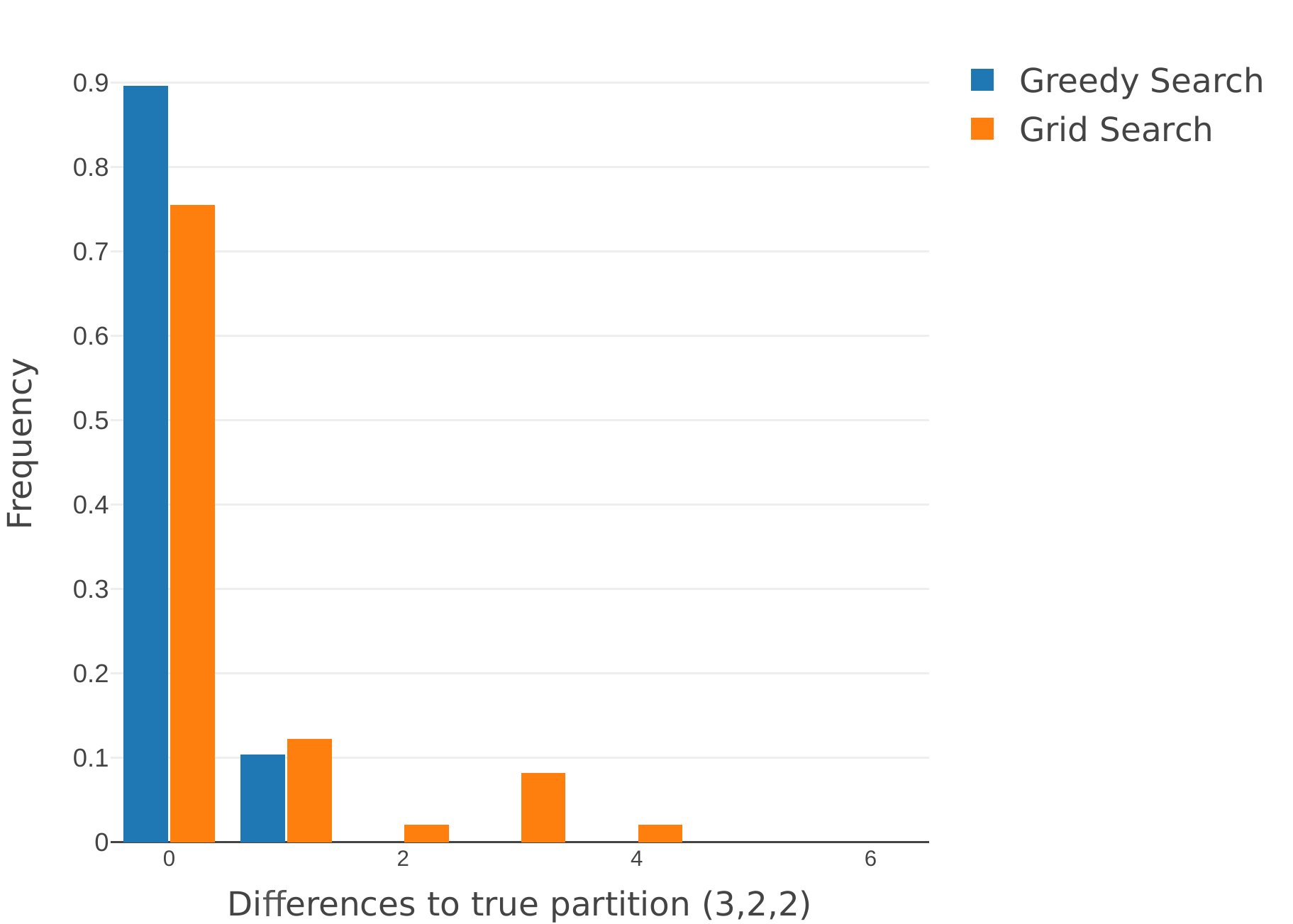}
	\caption{Results of 50 runs of each selection model strategy, in terms of differences to the generative model structure.}
	\label{fig:modelselectionaccuracy}
\end{figure}

The results of 50 launches of each method is displayed in Fig.~\ref{fig:modelselectionaccuracy}, in terms of differences to the true partition. While the grid search superiority was predictable, the results illustrate the differences in results, to be compared to the computation resources required. From our perspective, the grid search approach seems better, as long as we keep the clusters number low.

\section{Application to autonomous driving system validation}
The Scala/Spark source code of the method is available at the github repository https://github.com/EtienneGof/FunCLBM, along with the data simulation method. The real-case data is not, however, put at disposal.

Validating an intelligent driving system is a complicated task, that can not be purely addressed with on-track tests. The numerical simulation approach circumvents the limits of these physical experiments, mainly due to the high numbers of validation check to perform to assess a system. A large scale simulation framework reproducing test conditions is intensively used to test driverless cars, producing a massive amount of time series that needs to be processed.
Several aspects motivate the use of an autonomous behavior simulation platform. One of the main motivation is the physical validation cost reduction. Such validations require specific infrastructures, equipment management and maintenance, and significant human intervention to set up the experiments.

Another major disadvantage of physical testing is the impossibility to produce enough sample to prove the high reliability of a system. A validation objective may be the assessment of vehicle incident odds (e.g. $<10^{-8}$ incidents per hour). With a classical sampling method, estimating such probability would require running prototypes over hundreds of millions of kilometers.
Therefore, using a digital environment to test the different vehicles enables us to reproduce an exact experimental setting, to repeat the tests on-demand in an automated fashion in parallel of the development of the control software, as well as sample the test input parameters to assess the uncertainty of the experiments and the robustness of the cars.
Even if such a large amount of real-life data were available, as is the case in some data science applications, there would be no guarantees about neither the data quality nor value. In our case, this value lies in the specific driving situation in which to test the control logic reaction. These situations might be rarely occurring in real-life driving sessions, such as emergency braking or lane departure events.

Therefore, the use of an high-perfomance computing environment provides a mean to extensively test a driving control logic. Because of the large variety and complexity of the simulated driving situations, as well as the \emph{possibly unknown} operating modes of the intelligent car, using a supervised approach is intractable for the massive datasets under consideration.

\subsection{Use case description}

\begin{figure}[h!]
	\centering
	\includegraphics[width=0.7\linewidth]{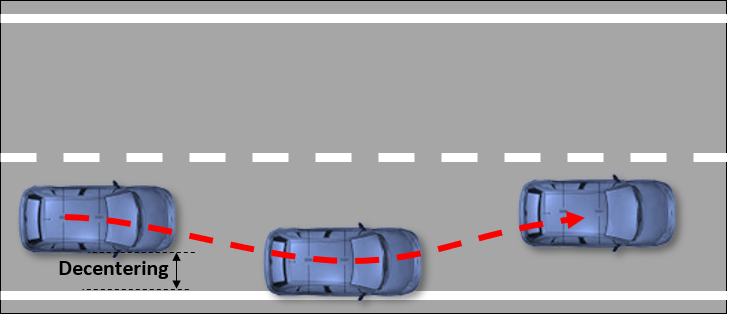}
	\caption{Use case illustration: Ego drifts from the runway's center line and cross the white line on the side of the road, before being put back in the runway center}
	\label{fig:visuUseCase1}
\end{figure}

In this situation, the objective is to test the reactions of a car (called Ego) equipped with the control logic. Ego runs in a straight line and starts drifting laterally towards the road side or the other lane, simulating a sleeping driver. We expect the drifting detection system to trigger the control logic, which in turn puts the car back in its line center, as an emergency maneuver. The situation is depicted in Fig.~\ref{fig:visuUseCase1}.

The simulated datasets contain the data from 56 simulations, each described by 20 signals. Some signals are duplicated in order to test FunCLBM ability to regroup them, and some uninformative ones are kept on purpose.

\subsection{Results}

The experiments on simulated datasets lead us to choose the following setup for the real case analysis: initialization is always performed by sampling column and row cluster partitions, and the FunLBM grid search approach is applied for model selection. Each combination is tested with 30 concurrent runs. 

\begin{figure}[h!]
	\centering
	\includegraphics[width=0.7\linewidth]{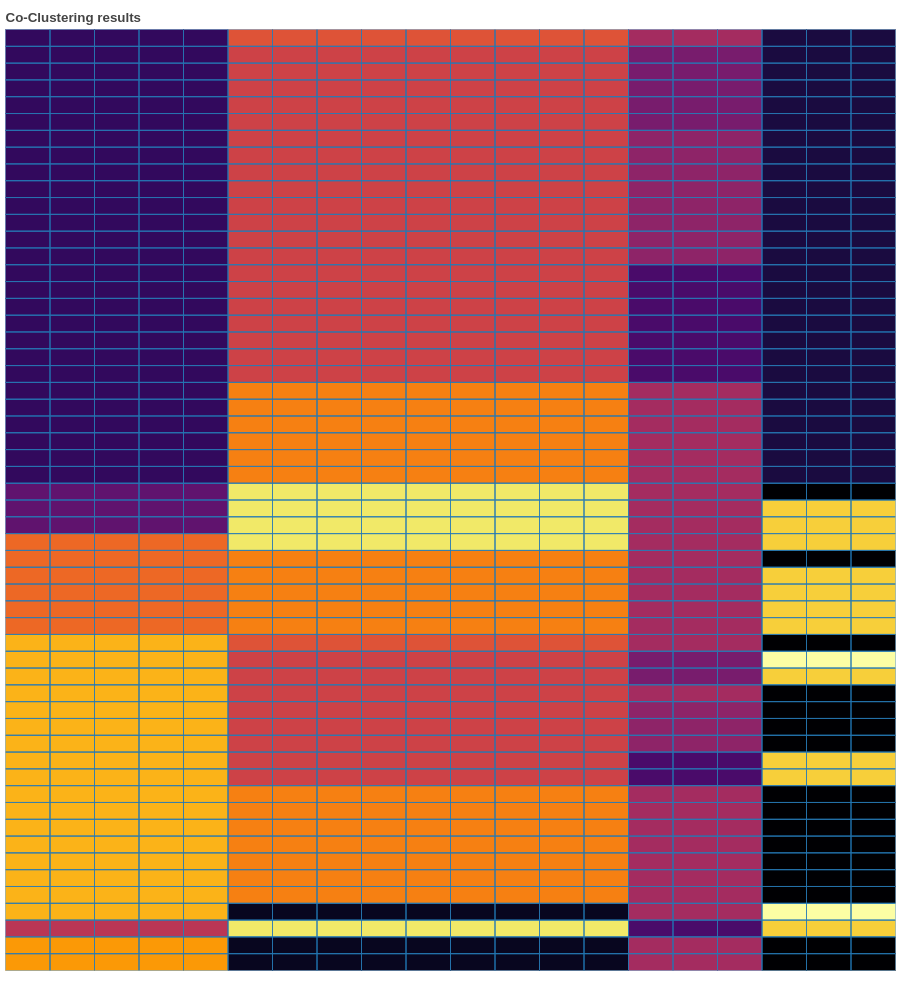}
	\caption{Final structure obtained on real case dataset}
	\label{fig:realCaseCoclustering}
\end{figure}

The final clustering structure is presented in Fig.\ref{fig:realCaseCoclustering}. It consists of 4 column clusters, each one with a different number of row clusters: (6x5x4x4). Due to constraints on article length, each of these 19 clusters cannot be analyzed in-depth here. However we give the main insights below. 

The first column cluster groups the following features: Ego's current lateral lane position (continuous), Ego's current lane index (discrete), type of the lane on Ego's right side (discrete), and type of lane on Ego's left side (discrete). The last two signals seem to be wrongly clustered at first sight, but are in fact redundant Ego's position, as they uniquely identify Ego's current lane index. Interestingly, this first column cluster therefore gathers every features related to the position of Ego. 

\begin{figure}
	\centering
	\includegraphics[width=0.6\linewidth]{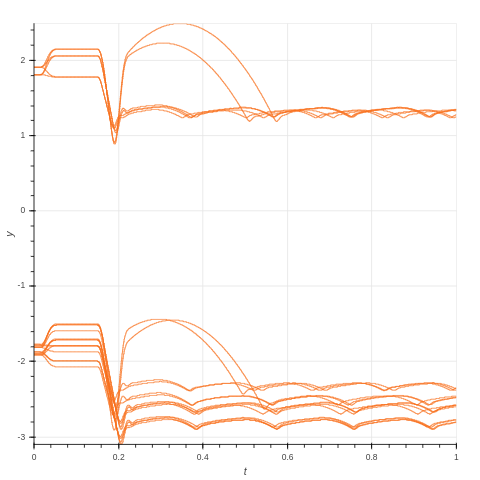}
	\caption{Ego lateral position in Block Cluster (5,1)}
	\label{fig:BC05}
\end{figure}

\begin{figure}
	\centering
	\includegraphics[width=0.6\linewidth]{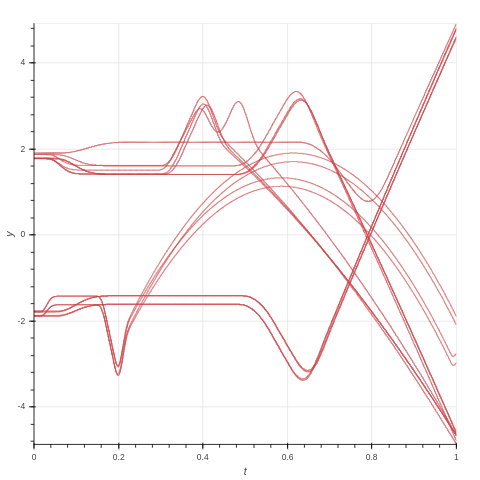}
	\caption{Ego lateral position in Block Cluster (2,1)}
	\label{fig:BC02}
\end{figure}

\begin{figure}
	\centering
	\includegraphics[width=0.6\linewidth]{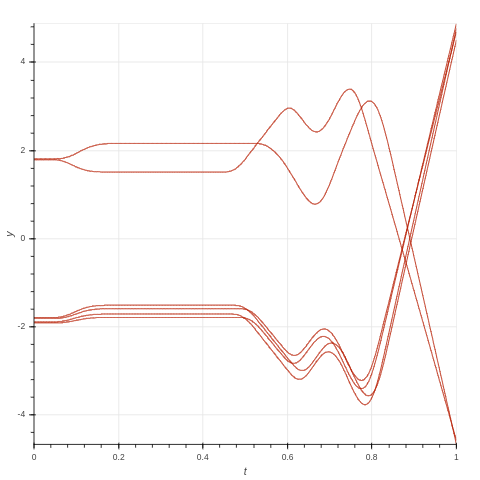}
	\caption{Ego lateral position in Block Cluster (3,1)}
	\label{fig:BC03}
\end{figure}

The conditional row partitioning in this column cluster is also interesting: the partition of Ego's position signal is represented in Figs. ~\ref{fig:BC05}, ~\ref{fig:BC02}. The clusters adequately gather simulations that share the same behavior. In Fig. ~\ref{fig:BC05} case, the control logic is activated and the car is recentered in its lane, and then repeatedly bounces back on the exact same road markings. In Fig. Fig~\ref{fig:BC03} case, the decentering happens later, and the car bounces once before changing direction and going straightforwardly to the other side of the road. In Fig~\ref{fig:BC03} scenario, the car bounces only once and either goes to other side of the road of comes back after a large drift.

\begin{figure}
	\centering
	\includegraphics[width=0.6\linewidth]{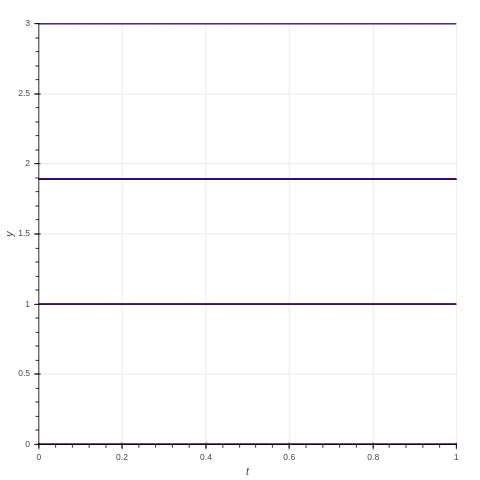}
	\caption{Uninformative signals in Block Cluster (2,2):  linearly increasing feature (vehicle's width, length, headlights activation..) }
	\label{fig:BC10}
\end{figure}

\begin{figure}
	\centering
	\includegraphics[width=0.6\linewidth]{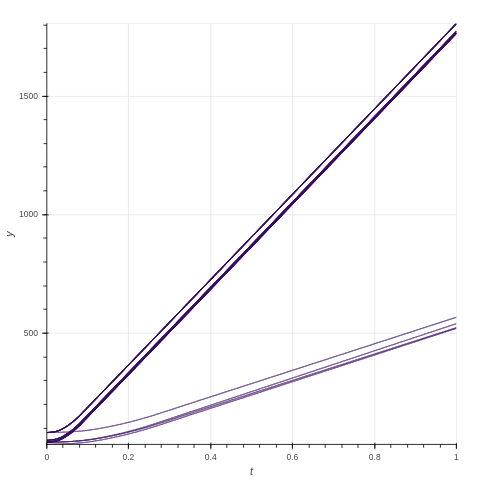}
	\caption{Uninformative signals in Block Cluster (1,2):  linearly increasing feature (distance to origin)}
	\label{fig:BC10_2}
\end{figure}

In the second, and largest, column-cluster can be found the uninformative signals, that either give constant values (vehicle length, width, distance between wheels, road bend radius) or increasing linearly (distance to origin). Fig.~\ref{fig:BC10} and  Fig.~\ref{fig:BC10_2} illustrates some of these signals.

\begin{figure}
	\centering
	\includegraphics[width=0.6\linewidth]{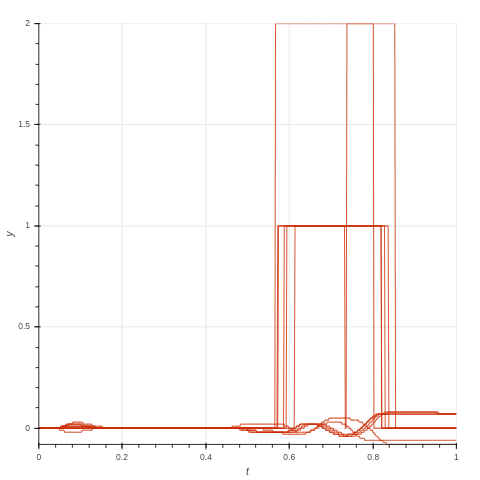}
	\caption{Control logic activation and changes in Ego's heading in Block Cluster (3,3)}
	\label{fig:BC22}
\end{figure}

The third column-cluster regroups two other interesting features: the rectangular function indicating the activation of the control logic, and the changes in Ego's heading. While the first column-cluster was grouping position, this one gathers the control features. Fig.\ref{fig:BC22} shows the content of subcluster (3, 3) which illustrates the relationship between them. Overall, every set of duplicate features have been correctly grouped together.

This conditional clustering partition shows, in conclusion, that the FunCLBM approach has correctly discriminated uninformative signals, while creating meaningful clusters of features (position and leverage). In each column-clusters, the observatins are also informative and provide good insights of the dataset content.

\section{Conclusions and Future Work}

This paper describes FunCLBM, a model-based method which addresses the problem of clustering multivariate time series in multi-views. This new model enables regrouping redundant signals, discriminating uninformative ones and provides the user with multiple clustering views of a multivariate time series dataset.

The time series are transformed into interpolated log-periodogram before being projected into low-dimensional space. This space is adapted to each block-cluster, and updated at each iteration of a SEM-Gibbs algorithm for model inference.

Several initialization methods and model selection strategies are proposed and experimented on a simulated dataset, which shows the model adequacy and give insights on the most interesting implementation strategies. Finally, we apply the method to a real-case dataset from the autonomous driving system validation domain. In this application, FunCLBM has been able to simultaneously discriminate groups of signals and produce meaningful driving behavior clusters. These results shows the usefulness of the model and the effectiveness of the initialization and model selection strategy. 

The FunCLBM approach was applied here to an autonomous driving context, however we are confident that it can be used in many other domains. Several improvements are being considered in order to facilitate its use. The model selection, for instance, can become computationally expensive  when the number of observations and signals increases. Similarly, the initialization can become problematic for higher numbers of clusters. In order to overcome these constraints, we plan to investigate new initialization methods based on Importance Sampling, as well as new model selection strategies. In this context, the development of a non-parametric functional conditional latent block model seems a promising lead. 

\end{sloppypar}

\bibliographystyle{natbib}

\begin{thebibliography}{20}
	
	%%% the Web address.
	
	\ifx \showCODEN    \undefined \def \showCODEN     #1{\unskip}     \fi
	\ifx \showDOI      \undefined \def \showDOI       #1{#1}\fi
	\ifx \showISBNx    \undefined \def \showISBNx     #1{\unskip}     \fi
	\ifx \showISBNxiii \undefined \def \showISBNxiii  #1{\unskip}     \fi
	\ifx \showISSN     \undefined \def \showISSN      #1{\unskip}     \fi
	\ifx \showLCCN     \undefined \def \showLCCN      #1{\unskip}     \fi
	\ifx \shownote     \undefined \def \shownote      #1{#1}          \fi
	\ifx \showarticletitle \undefined \def \showarticletitle #1{#1}   \fi
	\ifx \showURL      \undefined \def \showURL       {\relax}        \fi
	% invisible to TeX
	\providecommand\bibfield[2]{#2}
	\providecommand\bibinfo[2]{#2}
	\providecommand\natexlab[1]{#1}
	\providecommand\showeprint[2][]{arXiv:#2}
	
	\bibitem[\protect\citeauthoryear{Aghabozorgi, Shirkhorshidi, and
		Wah}{Aghabozorgi et~al\mbox{.}}{2015}]%
	{aghabozorgi2015time}
	\bibfield{author}{\bibinfo{person}{Saeed Aghabozorgi},
		\bibinfo{person}{Ali~Seyed Shirkhorshidi}, {and} \bibinfo{person}{Teh~Ying
			Wah}.} \bibinfo{year}{2015}\natexlab{}.
	\newblock \showarticletitle{Time-series clustering--a decade review}.
	\newblock \bibinfo{journal}{\emph{Information Systems}}  \bibinfo{volume}{53}
	(\bibinfo{year}{2015}), \bibinfo{pages}{16--38}.
	\newblock
	
	
	\bibitem[\protect\citeauthoryear{Bagnall, Lines, Bostrom, Large, and
		Keogh}{Bagnall et~al\mbox{.}}{2017}]%
	{bagnall2017great}
	\bibfield{author}{\bibinfo{person}{Anthony Bagnall}, \bibinfo{person}{Jason
			Lines}, \bibinfo{person}{Aaron Bostrom}, \bibinfo{person}{James Large}, {and}
		\bibinfo{person}{Eamonn Keogh}.} \bibinfo{year}{2017}\natexlab{}.
	\newblock \showarticletitle{The great time series classification bake off: a
		review and experimental evaluation of recent algorithmic advances}.
	\newblock \bibinfo{journal}{\emph{Data Mining and Knowledge Discovery}}
	\bibinfo{volume}{31}, \bibinfo{number}{3} (\bibinfo{year}{2017}),
	\bibinfo{pages}{606--660}.
	\newblock
	
	
	\bibitem[\protect\citeauthoryear{Baudry and Celeux}{Baudry and Celeux}{2015}]%
	{baudry2015mixtures}
	\bibfield{author}{\bibinfo{person}{Jean-Patrick Baudry} {and}
		\bibinfo{person}{Gilles Celeux}.} \bibinfo{year}{2015}\natexlab{}.
	\newblock \showarticletitle{EM for mixtures}.
	\newblock \bibinfo{journal}{\emph{Statistics and computing}}
	\bibinfo{volume}{25}, \bibinfo{number}{4} (\bibinfo{year}{2015}),
	\bibinfo{pages}{713--726}.
	\newblock
	
	
	\bibitem[\protect\citeauthoryear{Biernacki, Celeux, and Govaert}{Biernacki
		et~al\mbox{.}}{2000}]%
	{biernacki2000assessing}
	\bibfield{author}{\bibinfo{person}{Christophe Biernacki},
		\bibinfo{person}{Gilles Celeux}, {and} \bibinfo{person}{G{\'e}rard Govaert}.}
	\bibinfo{year}{2000}\natexlab{}.
	\newblock \showarticletitle{Assessing a mixture model for clustering with the
		integrated completed likelihood}.
	\newblock \bibinfo{journal}{\emph{IEEE transactions on pattern analysis and
			machine intelligence}} \bibinfo{volume}{22}, \bibinfo{number}{7}
	(\bibinfo{year}{2000}), \bibinfo{pages}{719--725}.
	\newblock
	
	
	\bibitem[\protect\citeauthoryear{Biernacki, Celeux, and Govaert}{Biernacki
		et~al\mbox{.}}{2003}]%
	{biernacki2003choosing}
	\bibfield{author}{\bibinfo{person}{Christophe Biernacki},
		\bibinfo{person}{Gilles Celeux}, {and} \bibinfo{person}{G{\'e}rard Govaert}.}
	\bibinfo{year}{2003}\natexlab{}.
	\newblock \showarticletitle{Choosing starting values for the EM algorithm for
		getting the highest likelihood in multivariate Gaussian mixture models}.
	\newblock \bibinfo{journal}{\emph{Computational Statistics \& Data Analysis}}
	\bibinfo{volume}{41}, \bibinfo{number}{3-4} (\bibinfo{year}{2003}),
	\bibinfo{pages}{561--575}.
	\newblock
	
	
	\bibitem[\protect\citeauthoryear{Bl{\"o}mer and Bujna}{Bl{\"o}mer and
		Bujna}{2013}]%
	{blomer2013simple}
	\bibfield{author}{\bibinfo{person}{Johannes Bl{\"o}mer} {and}
		\bibinfo{person}{Kathrin Bujna}.} \bibinfo{year}{2013}\natexlab{}.
	\newblock \showarticletitle{Simple methods for initializing the em algorithm
		for gaussian mixture models}.
	\newblock \bibinfo{journal}{\emph{CoRR}} (\bibinfo{year}{2013}).
	\newblock
	
	
	\bibitem[\protect\citeauthoryear{Bouveyron, Bozzi, Jacques, and
		Jollois}{Bouveyron et~al\mbox{.}}{2018}]%
	{bouveyron2018functional}
	\bibfield{author}{\bibinfo{person}{Charles Bouveyron}, \bibinfo{person}{Laurent
			Bozzi}, \bibinfo{person}{Julien Jacques}, {and}
		\bibinfo{person}{Fran{\c{c}}ois-Xavier Jollois}.}
	\bibinfo{year}{2018}\natexlab{}.
	\newblock \showarticletitle{The functional latent block model for the
		co-clustering of electricity consumption curves}.
	\newblock \bibinfo{journal}{\emph{Journal of the Royal Statistical Society:
			Series C (Applied Statistics)}} \bibinfo{volume}{67}, \bibinfo{number}{4}
	(\bibinfo{year}{2018}), \bibinfo{pages}{897--915}.
	\newblock
	
	
	\bibitem[\protect\citeauthoryear{Bouveyron and Jacques}{Bouveyron and
		Jacques}{2011}]%
	{bouveyron2011model}
	\bibfield{author}{\bibinfo{person}{Charles Bouveyron} {and}
		\bibinfo{person}{Julien Jacques}.} \bibinfo{year}{2011}\natexlab{}.
	\newblock \showarticletitle{Model-based clustering of time series in
		group-specific functional subspaces}.
	\newblock \bibinfo{journal}{\emph{Advances in Data Analysis and
			Classification}} \bibinfo{volume}{5}, \bibinfo{number}{4}
	(\bibinfo{year}{2011}), \bibinfo{pages}{281--300}.
	\newblock
	
	
	\bibitem[\protect\citeauthoryear{Caiado, Crato, and Pe{\~n}a}{Caiado
		et~al\mbox{.}}{2009}]%
	{caiado2009comparison}
	\bibfield{author}{\bibinfo{person}{Jorge Caiado}, \bibinfo{person}{Nuno Crato},
		{and} \bibinfo{person}{Daniel Pe{\~n}a}.} \bibinfo{year}{2009}\natexlab{}.
	\newblock \showarticletitle{Comparison of times series with unequal length in
		the frequency domain}.
	\newblock \bibinfo{journal}{\emph{Communications in Statistics—Simulation and
			Computation{\textregistered}}} \bibinfo{volume}{38}, \bibinfo{number}{3}
	(\bibinfo{year}{2009}), \bibinfo{pages}{527--540}.
	\newblock
	
	
	\bibitem[\protect\citeauthoryear{Chamroukhi and Biernacki}{Chamroukhi and
		Biernacki}{2017}]%
	{chamroukhi2017model}
	\bibfield{author}{\bibinfo{person}{Faicel Chamroukhi} {and}
		\bibinfo{person}{Christophe Biernacki}.} \bibinfo{year}{2017}\natexlab{}.
	\newblock \showarticletitle{Model-Based Co-Clustering of Multivariate
		Functional Data}.
	\newblock
	
	
	\bibitem[\protect\citeauthoryear{Chamroukhi, Sam{\'e}, Govaert, and
		Aknin}{Chamroukhi et~al\mbox{.}}{2010}]%
	{chamroukhi2010hidden}
	\bibfield{author}{\bibinfo{person}{Faicel Chamroukhi}, \bibinfo{person}{Allou
			Sam{\'e}}, \bibinfo{person}{G{\'e}rard Govaert}, {and}
		\bibinfo{person}{Patrice Aknin}.} \bibinfo{year}{2010}\natexlab{}.
	\newblock \showarticletitle{A hidden process regression model for functional
		data description. application to curve discrimination}.
	\newblock \bibinfo{journal}{\emph{Neurocomputing}} \bibinfo{volume}{73},
	\bibinfo{number}{7-9} (\bibinfo{year}{2010}), \bibinfo{pages}{1210--1221}.
	\newblock
	
	
	\bibitem[\protect\citeauthoryear{Dempster, Laird, and Rubin}{Dempster
		et~al\mbox{.}}{1977}]%
	{dempster1977maximum}
	\bibfield{author}{\bibinfo{person}{Arthur~P Dempster}, \bibinfo{person}{Nan~M
			Laird}, {and} \bibinfo{person}{Donald~B Rubin}.}
	\bibinfo{year}{1977}\natexlab{}.
	\newblock \showarticletitle{Maximum likelihood from incomplete data via the EM
		algorithm}.
	\newblock \bibinfo{journal}{\emph{Journal of the Royal Statistical Society:
			Series B (Methodological)}} \bibinfo{volume}{39}, \bibinfo{number}{1}
	(\bibinfo{year}{1977}), \bibinfo{pages}{1--22}.
	\newblock
	
	
	\bibitem[\protect\citeauthoryear{Dhillon}{Dhillon}{2001}]%
	{dhillon2001co}
	\bibfield{author}{\bibinfo{person}{Inderjit~S Dhillon}.}
	\bibinfo{year}{2001}\natexlab{}.
	\newblock \showarticletitle{Co-clustering documents and words using bipartite
		spectral graph partitioning}. In \bibinfo{booktitle}{\emph{Proceedings of the
			seventh ACM SIGKDD international conference on Knowledge discovery and data
			mining}}. \bibinfo{pages}{269--274}.
	\newblock
	
	
	\bibitem[\protect\citeauthoryear{Govaert and Nadif}{Govaert and Nadif}{2003}]%
	{govaert2003clustering}
	\bibfield{author}{\bibinfo{person}{G{\'e}rard Govaert} {and}
		\bibinfo{person}{Mohamed Nadif}.} \bibinfo{year}{2003}\natexlab{}.
	\newblock \showarticletitle{Clustering with block mixture models}.
	\newblock \bibinfo{journal}{\emph{Pattern Recognition}} \bibinfo{volume}{36},
	\bibinfo{number}{2} (\bibinfo{year}{2003}), \bibinfo{pages}{463--473}.
	\newblock
	
	
	\bibitem[\protect\citeauthoryear{Govaert and Nadif}{Govaert and Nadif}{2013}]%
	{govaert2013co}
	\bibfield{author}{\bibinfo{person}{G{\'e}rard Govaert} {and}
		\bibinfo{person}{Mohamed Nadif}.} \bibinfo{year}{2013}\natexlab{}.
	\newblock \bibinfo{booktitle}{\emph{Co-clustering: models, algorithms and
			applications}}.
	\newblock \bibinfo{publisher}{John Wiley \& Sons}.
	\newblock
	
	
	\bibitem[\protect\citeauthoryear{Jacques and Biernacki}{Jacques and
		Biernacki}{2018}]%
	{jacques2018model}
	\bibfield{author}{\bibinfo{person}{Julien Jacques} {and}
		\bibinfo{person}{Christophe Biernacki}.} \bibinfo{year}{2018}\natexlab{}.
	\newblock \showarticletitle{Model-based co-clustering for ordinal data}.
	\newblock \bibinfo{journal}{\emph{Computational Statistics \& Data Analysis}}
	\bibinfo{volume}{123} (\bibinfo{year}{2018}), \bibinfo{pages}{101--115}.
	\newblock
	
	
	\bibitem[\protect\citeauthoryear{Keribin, Govaert, and Celeux}{Keribin
		et~al\mbox{.}}{2010}]%
	{keribin2010estimation}
	\bibfield{author}{\bibinfo{person}{Christine Keribin},
		\bibinfo{person}{G{\'e}rard Govaert}, {and} \bibinfo{person}{Gilles Celeux}.}
	\bibinfo{year}{2010}\natexlab{}.
	\newblock \showarticletitle{Estimation d'un mod{\`e}le {\`a} blocs latents par
		l'algorithme SEM}.
	\newblock
	
	
	\bibitem[\protect\citeauthoryear{Lomet}{Lomet}{2012}]%
	{lomet2012selection}
	\bibfield{author}{\bibinfo{person}{Aurore Lomet}.}
	\bibinfo{year}{2012}\natexlab{}.
	\newblock \emph{\bibinfo{title}{S{\'e}lection de mod{\`e}le pour la
			classification crois{\'e}e de don\-nées continues}}.
	\newblock \bibinfo{thesistype}{Ph.D. Dissertation}.
	\bibinfo{school}{Compi{\`e}gne}.
	\newblock
	
	\bibitem[\protect\citeauthoryear{Marbac, and Vandewalle}{Marbac
		et~al\mbox{.}}{2019}]%
	{marbac2019tractable}
	\bibfield{author}{\bibinfo{person}{Matthieu Marbac}, \bibinfo{person}{Vincent Vandewalle}} \bibinfo{year}{2019}\natexlab{}.
	\newblock \showarticletitle{A tractable multi-partitions clustering}. In \bibinfo{booktitle}{\emph{Computational Statistics \& Data Analysis}}. \bibinfo{pages}{167--179}.
	\newblock
	
	\bibitem[\protect\citeauthoryear{Ramsay and Silverman}{Ramsay and
		Silverman}{2005}]%
	{ramsay2005principal}
	\bibfield{author}{\bibinfo{person}{JO Ramsay} {and} \bibinfo{person}{BW
			Silverman}.} \bibinfo{year}{2005}\natexlab{}.
	\newblock \showarticletitle{Principal components analysis for functional data}.
	\newblock \bibinfo{journal}{\emph{Functional data analysis}}
	(\bibinfo{year}{2005}), \bibinfo{pages}{147--172}.
	\newblock
	
	
	\bibitem[\protect\citeauthoryear{Schmutz, Jacques, Bouveyron, Ch{\`e}ze, and
		Martin}{Schmutz et~al\mbox{.}}{2019}]%
	{schmutz2019co}
	\bibfield{author}{\bibinfo{person}{Amandine Schmutz}, \bibinfo{person}{Julien
			Jacques}, \bibinfo{person}{Charles Bouveyron}, \bibinfo{person}{Laurence
			Ch{\`e}ze}, {and} \bibinfo{person}{Pauline Martin}.}
	\bibinfo{year}{2019}\natexlab{}.
	\newblock \showarticletitle{Co-clustering de courbes fonctionnelles
		multivari{\'e}es}.
	\newblock
	
	\bibitem[\protect\citeauthoryear{Slimen, Allio, and Jacques}{Slimen
		et~al\mbox{.}}{2018}]%
	{slimen2018model}
	\bibfield{author}{\bibinfo{person}{Yosra~Ben Slimen}, \bibinfo{person}{Sylvain
			Allio}, {and} \bibinfo{person}{Julien Jacques}.}
	\bibinfo{year}{2018}\natexlab{}.
	\newblock \showarticletitle{Model-based co-clustering for functional data}.
	\newblock \bibinfo{journal}{\emph{Neurocomputing}}  \bibinfo{volume}{291}
	(\bibinfo{year}{2018}), \bibinfo{pages}{97--108}.
	\newblock
	
	
	\bibitem[\protect\citeauthoryear{Vlassis and Likas}{Vlassis and Likas}{2002}]%
	{vlassis2002greedy}
	\bibfield{author}{\bibinfo{person}{Nikos Vlassis} {and}
		\bibinfo{person}{Aristidis Likas}.} \bibinfo{year}{2002}\natexlab{}.
	\newblock \showarticletitle{A greedy EM algorithm for Gaussian mixture
		learning}.
	\newblock \bibinfo{journal}{\emph{Neural processing letters}}
	\bibinfo{volume}{15}, \bibinfo{number}{1} (\bibinfo{year}{2002}),
	\bibinfo{pages}{77--87}.
	\newblock
	
	
	\bibitem[\protect\citeauthoryear{Xu, Cheng, Zong, Ni, Song, Yu, Chen, Chen, and
		Zhang}{Xu et~al\mbox{.}}{2019}]%
	{xu2019deep}
	\bibfield{author}{\bibinfo{person}{Dongkuan Xu}, \bibinfo{person}{Wei Cheng},
		\bibinfo{person}{Bo Zong}, \bibinfo{person}{Jingchao Ni},
		\bibinfo{person}{Dongjin Song}, \bibinfo{person}{Wenchao Yu},
		\bibinfo{person}{Yuncong Chen}, \bibinfo{person}{Haifeng Chen}, {and}
		\bibinfo{person}{Xiang Zhang}.} \bibinfo{year}{2019}\natexlab{}.
	\newblock \showarticletitle{Deep co-clustering}. In
	\bibinfo{booktitle}{\emph{Proceedings of the 2019 SIAM International
			Conference on Data Mining}}. SIAM, \bibinfo{pages}{414--422}.
	\newblock
	
	
	\bibitem[\protect\citeauthoryear{Yang, Deng, Zheng, Yan, and Liu}{Yang
		et~al\mbox{.}}{2019}]%
	{yang2019deep}
	\bibfield{author}{\bibinfo{person}{Xu Yang}, \bibinfo{person}{Cheng Deng},
		\bibinfo{person}{Feng Zheng}, \bibinfo{person}{Junchi Yan}, {and}
		\bibinfo{person}{Wei Liu}.} \bibinfo{year}{2019}\natexlab{}.
	\newblock \showarticletitle{Deep spectral clustering using dual autoencoder
		network}. In \bibinfo{booktitle}{\emph{Proceedings of the IEEE Conference on
			Computer Vision and Pattern Recognition}}. \bibinfo{pages}{4066--4075}.
	\newblock
	
	
\end{thebibliography}

\end{document}